%% file: root.tex
\documentclass[letterpaper, 10 pt, conference]{ieeeconf}  

\IEEEoverridecommandlockouts                              

\title{\LARGE \bf
Mesh-Learner: Texturing Mesh with Spherical Harmonics
}

\usepackage{graphicx}

\usepackage{gensymb}
\usepackage{amsmath}
\graphicspath{{images/}}

\usepackage{color}
\usepackage[normalem]{ulem}
\usepackage{tabularray}
\usepackage{multirow} 
\usepackage{multicol} 
\usepackage{arydshln}
\usepackage{caption}
\usepackage{soul}
\usepackage{booktabs}
\usepackage{hyperref} 
\usepackage{placeins}
\usepackage{balance} 
\usepackage{subfig}
\usepackage[linesnumbered, ruled]{algorithm2e}

\iffalse
\definecolor{orange}{rgb}{0.8,0.5,0.0} 

\newcommand{\HL}[1]{\textcolor{blue}{#1}}

\else

\newcommand{\HL}[1]{{#1}}

\fi

\author{Yunfei Wan, Jianheng Liu, Chunran Zheng, Jiarong Lin$^{*}$ and Fu Zhang$^{*}$
\thanks{Y. Wan, J. Lin, C. Zheng, J. Liu and F. Zhang are with the Department of Mechanical Engineering, The University of Hong Kong, Hong Kong SAR, China. {\tt\footnotesize $\{$alexwan, jianheng, zhengcr, zivlin$\}$@connect.hku.hk, fuzhang@hku.hk}}
\thanks{$^{*}$Corresponding author: Jiarong Lin and Fu Zhang}
}

\begin{document}

\maketitle
\thispagestyle{empty}
\pagestyle{empty}

\begin{abstract}

\HL{In this paper, we present a 3D reconstruction and rendering framework termed \textit{Mesh-Learner} that is natively compatible with traditional rasterization pipelines. It integrates mesh and spherical harmonic (SH) Texture (i.e., texture filled with SH coefficients) into the learning process to learn each mesh's view-dependent radiance end-to-end.} 
Images are rendered by interpolating surrounding \textit{SH Texels} at each pixel's sampling point using a novel interpolation method. Conversely, gradients from each pixel are back-propagated to the related \textit{SH Texels} in \textit{SH Textures}.
 \textit{Mesh-Learner} exploits graphic features of rasterization pipeline (texture sampling, deferred rendering) to render, which makes \textit{Mesh-Learner} naturally compatible with tools (e.g., Blender) and tasks (e.g., 3D reconstruction, scene rendering, reinforcement learning for robotics) that are based on rasterization pipelines.
Our system can train vast, unlimited scenes because we transfer only the \textit{SH Textures} within the frustum to the GPU for training. At other times, the \textit{SH Textures} are stored in CPU RAM, which results in moderate GPU memory usage.

The rendering results on interpolation and extrapolation sequences in the Replica and FAST-LIVO2 datasets achieve state-of-the-art performance compared to existing state-of-the-art methods (e.g., 3D Gaussian Splatting and M2-Mapping). To benefit the society, the code will be available at \url{https://github.com/hku-mars/Mesh-Learner}. 

\end{abstract}

\section{INTRODUCTION}
  Nowadays, 3D reconstruction and novel view synthesis are thriving research directions, as they can be applied in simulation task \cite{mittal2023orbit}, robot's navigation \cite{imap}, entertainment \cite{meta_XR_SDK} applications, and so on. 
  In the past, traditional algorithms like multi-view stereo (MVS) \cite{mvs_review} were commonly used in this field. Recently, researchers have turned to Neural Radiance Fields (NeRF) \cite{mildenhall2020nerf} to address the shortcomings of traditional 3D primitives such as point clouds and voxels, enabling the use of continuous implicit functions parameterized by neural networks. However, these methods suffer from the limitations of real-time rendering and extended training durations. Subsequently, 3D Gaussian Splatting (3DGS) \cite{kerbl3Dgaussians} was proposed, offering significant improvements in rendering quality and training speed compared to NeRF.
Existing works such as \cite{nerfpp, Yu2024MipSplatting, reduce_3dgs} have significantly improved the rendering quality and efficiency of NeRF and 3DGS compared to their original versions \cite{kerbl3Dgaussians, mildenhall2020nerf}. However, they ignore a practical issue: inability to directly integrate with traditional rasterization pipelines. To visualize NeRF or 3DGS in rendering engines like Unity, users typically introduce third-party plugins, such as \cite{3DGS_unity_addon}.
  \begin{figure}[htbp]
    \centering
    \includegraphics[width=1.0\linewidth]{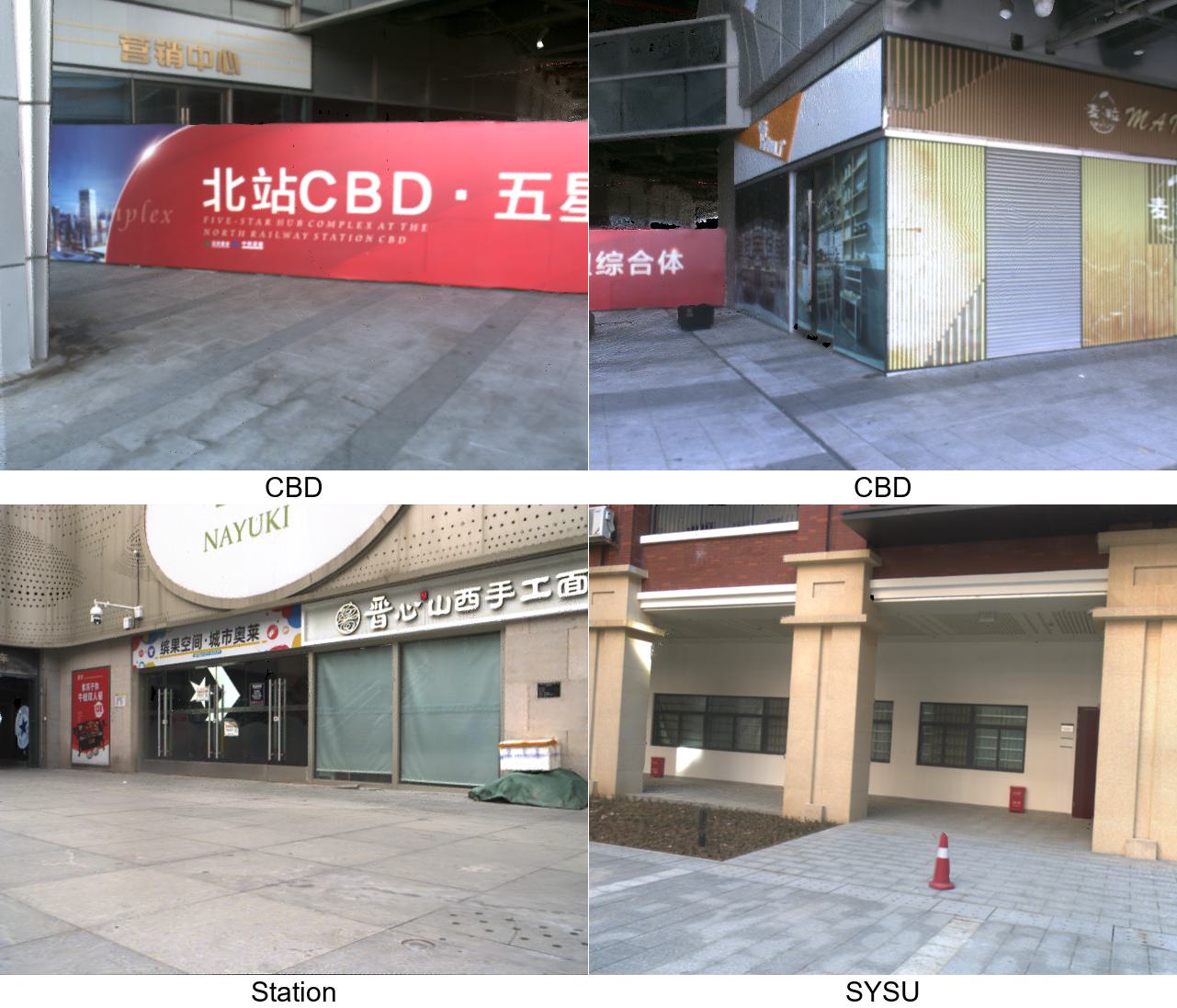}
    \caption{The rendering results of \textit{Mesh-learner} on real-world FAST-LIVO2 dataset.}
    \label{fig:first_page}
\end{figure}

  However, it suffers from artifacts \cite{unity_plugin_issue_artifacts} and fundamental rendering errors (e.g., occlusion error \cite{unity_plugin_issue_err_occlusion}). Furthermore, many applications already have extensive mesh models and require interaction with trained 3DGS or NeRF models.
  Therefore, there is a need to convert 3DGS or NeRF models into mesh format.
  However, these conversions not only introduce additional workload but also inevitably lead to losses in rendering quality and geometric integrity.
  
  Such limitations lead people to consider whether they can directly learn a texture from the existing mesh, so the learning outcome can be directly applied to a rasterization pipeline without conversion. This is based on the ability to obtain satisfactory meshes with the help of other lidar-based algorithms 
\cite{lin2023immesh, fast_livo2}.


  {To diminish mentioned gap between the algorithm outcome and the downstream tasks, we present \textit{Mesh-Learner}, an end-to-end framework based on rasterization pipeline for both training and inference, obtaining realistic novel view images.} \textit{Mesh-Learner} assigns each mesh a \textit{SH Texture}, utilizes rasterization pipeline to determine which \textit{SH Textures} are involved in rendering each frame and sample on them.
  At sample points, we propose a hybrid interpolation method between \textit{SH Texels} to obtain view-dependent colors. We extend the Elliptical Weighting Average (EWA) filter \cite{EWA_original} from local texture space into world space on \textit{SH Textures} to mitigate aliasing artifacts in distant scenes. In addition, we introduce an adaptive \textit{SH Density} strategy to determine the optimal \textit{SH Texture} resolution each mesh during training, according to the varying complexity of details in training scenes.

 \textit{Mesh-Learner} is developed in LibTorch C++ \cite{pytorch} and CUDA C \cite{cuda} with careful engineering optimization. Additionally, users of \textit{Mesh-Learner} can seamlessly integrate the training outcomes (\textit{SH Textures}) into their rasterization pipeline for visualization, as \textit{Mesh-Learner}'s inference (rendering) part can be implemented solely using OpenGL.
 In summary, the contributions of our work are:
  \begin{enumerate}
      \item A spherical-harmonic-augmented texture with a rasterization-pipeline-based training framework to enhance compatibility and capability of photorealistic rendering.
      \item A hybrid interpolation method that leverages adjacent \textit{SH Texture}'s information to represent view-dependent radiance.
      \item {A world space EWA filter to reduce aliasing for distant scenes.}
      \item {An adaptive \textit{SH Density} adjustment strategy to find the optimal \textit{SH Texture} resolution of each mesh.}
  \end{enumerate}

\section{RELATED WORKS}

Neural Radiance Fields (NeRF) \cite{mildenhall2020nerf} have pioneered a new track in 3D reconstruction and Novel View Synthesis (NVS) by utilizing neural networks to represent a scene. It maps a 3D position and a view direction to RGB color and density, employing volumetric rendering to produce photorealistic novel view images. Plenoxels \cite{plenoxels} attempts to model a scene without using neural networks; instead, it employs a sparse voxel grid to define a scene and place SH coefficients at each voxel. They utilize volume rendering to render. At each sample point, they calculate the color and opacity using trilinear interpolation of neighboring SHs, then integrate all sample points along the ray. Recently, 3DGS \cite{kerbl3Dgaussians} utilizes numerous Gaussian kernels instead of a fixed voxel grid to represent scene geometry and also uses SHs on each Gaussian kernel to store view-dependent radiance. They use rasterization to splat Gaussian kernels onto the image plane, sort them based on depth, and blend all the Gaussians at each pixel to composite the color. In comparison to them, the rendering method in our work combines rasterization and interpolation, aiming to achieve compatibility with the traditional rasterization pipeline and enhance rendering efficiency.

Several works introduce textures into 3DGS. \cite{chao2024texturedgaussiansenhanced3d} assign each Gaussian a distinctive texture. During rendering, they trace a ray from the camera and calculate the intersection point of the hit Gaussians. Then, they query the RGBA texture of each intersected Gaussian and composite the color and alpha to obtain the final color. Another similar work is \cite{song2024hdgstextured2dgaussian}. They introduce learnable texture maps on 2D Gaussian surfels, based on \cite{Huang2DGS2024}, enabling each primitive to enhance its ability to represent more complex and detailed appearances. They perform ray-surface intersection and calculate the texture coordinates for each surfel's texture, then obtain the zeroth basis of Spherical Harmonics by bilinearly sampling the texture map. 
In summary, the above methods treat texture as a medium attached to Gaussians to store colors. Unlike them, we directly implement rendering and optimization on textures.

Few works integrate traditional rendering pipelines into the training and rendering process of NeRF or 3DGS. \cite{chen2022mobilenerf} first represented the NeRF model based on textured polygons. They exploit the rasterization pipeline with Z-buffer to achieve pixel-level parallelism, providing each pixel with neural features in texture form. Then, they implement a MLP in the fragment shader to map those features to colors. However, this method is not end-to-end. First, they need to train a NeRF with continuous opacity, then binarize it, and finally bake opacity and features into textures for rendering. In comparison, our method is end-to-end and prioritizes ``high compatibility with traditional pipelines'' as the primary objective. Once the training converges, the output (\textit{SH Textures}) can be imported with the corresponding meshes into a rasterization pipeline for rendering, eliminating the need for any intermediate processing.

\section{METHODOLOGY}

\begin{figure}[htbp]
    \centering
    \includegraphics[width=1.0\linewidth]{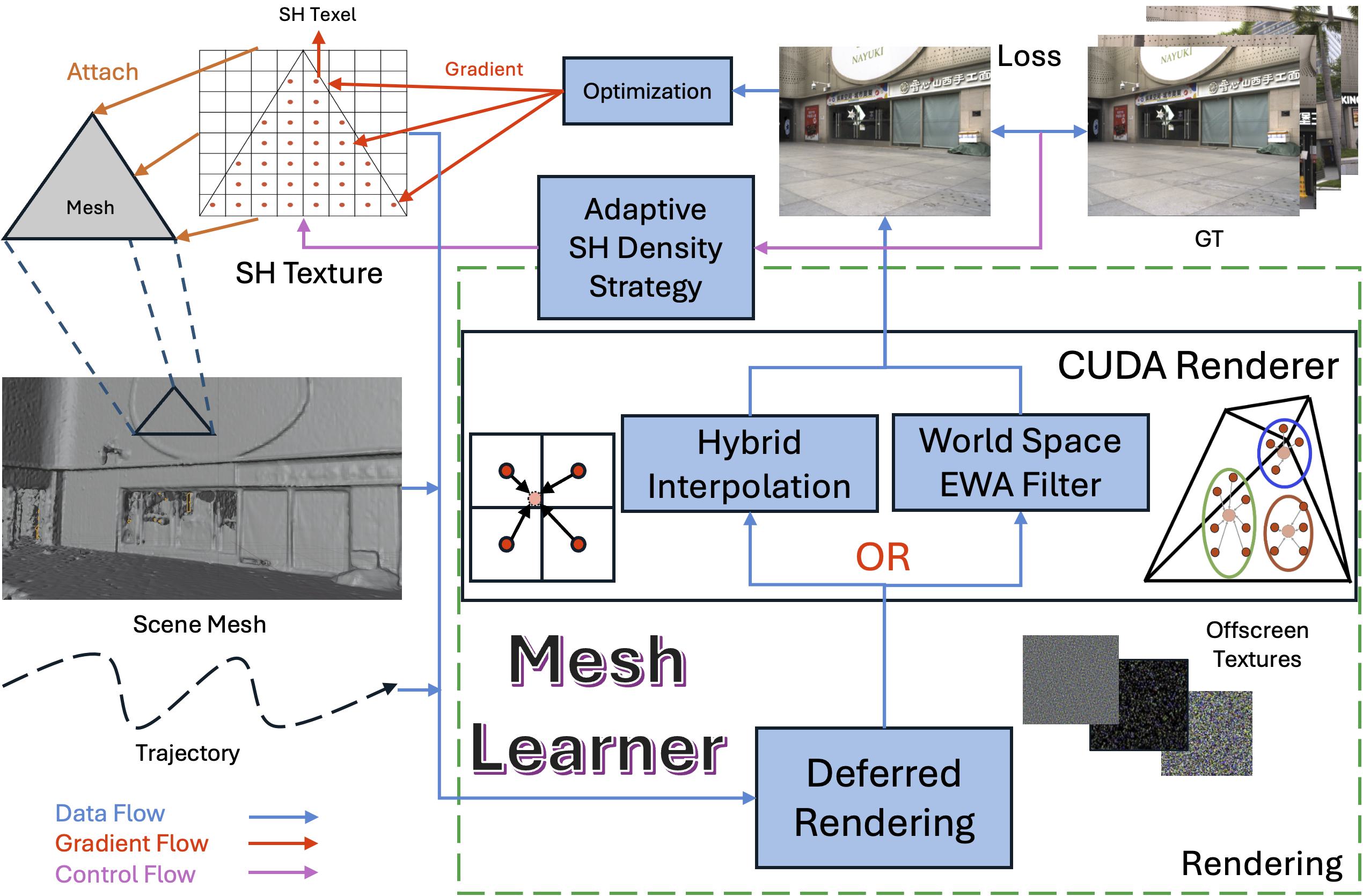}
    \caption{The overview of our proposed system.}
    \label{fig:sys_overview}
\end{figure}
The overview of our proposed system is illustrated in Fig. \ref{fig:sys_overview}. The goal is to attain each mesh's \textit{SH Texture} capable of rendering realistic view-dependent radiance. \HL{The system inputs are scene mesh and sampled posed images.} We obtain high-precision meshes by using existing lidar-based methods (e.g., ImMesh \cite{lin2023immesh}, FAST-LIVO2 \cite{fast_livo2} and M2-Mapping \cite{m2-mapping}). To initialize, we allocate a \textit{SH Texture} for each mesh whose resolution is calculated based on \textit{SH Density} (in Sec. \ref{III.A}). Then in rendering (forward), it is comprised of two parts:\begin{itemize}
    \item \textbf{Deferred rendering} ({in Sec. \ref{III.B})}: we rasterize meshes into screen space, save at each pixel's sampling texture coordinates, view directions, and other information of meshes in the frustum into offscreen textures.
    \item \textbf{CUDA Renderer} ({in Sec. \ref{III.C}}): we read offscreen textures from the deferred rendering stage, then on each pixel's sampling point on \textit{SH Textures}, we implement hybrid interpolation (in Sec. \ref{III.D}) between neighbor \textit{SH Texels} to get pixel colors. If a rendering view is distant, we instead utilize our proposed world space EWA filter (in Sec. \ref{III.E}) on \textit{SH Textures} to obtain anti-aliasing results.
\end{itemize} 
  In backward, a CUDA kernel is employed to record which \textit{SH Texels} contribute to each output pixel, whether through hybrid interpolation or world space EWA. Subsequently, based on the recorded relationships, the gradient of each pixel is backpropagated to the \textit{SH Texels} that influence the color of that pixel. Throughout the training, we periodically employ an adaptive \textit{SH Density} strategy (in Sec. \ref{III.F}) for each mesh to determine the optimal SH Texture resolution for each.

\subsection{SH Density}
\label{III.A}

Enlightened by Plenoxels \cite{plenoxels} that interpolates SHs in 3D grids, we interpolate SHs on 2D \textit{SH Textures}. We regard \textit{SH Texture} as a tangible entity attached to the mesh surface, thereby we can assign each texel center a specific world coordinate. A simple approach to rendering is to assign a uniform resolution config to all \textit{SH Textures} and implement bilinear interpolation at pixels' sampling points. However, this can lead to suboptimal rendering quality, as it overlooks that each mesh varies in size and shape (obtuse, acute, or right triangles). Specifically, in areas with high-frequency details, smaller meshes may yield satisfied interpolation results. However, on larger meshes with the same texture resolution config, \textit{SH Texels} might be too sparse to accurately represent those high-frequency details.

\begin{figure}[htbp]
	\centering
    

    \subfloat[]{
        \includegraphics[width=0.5\linewidth]{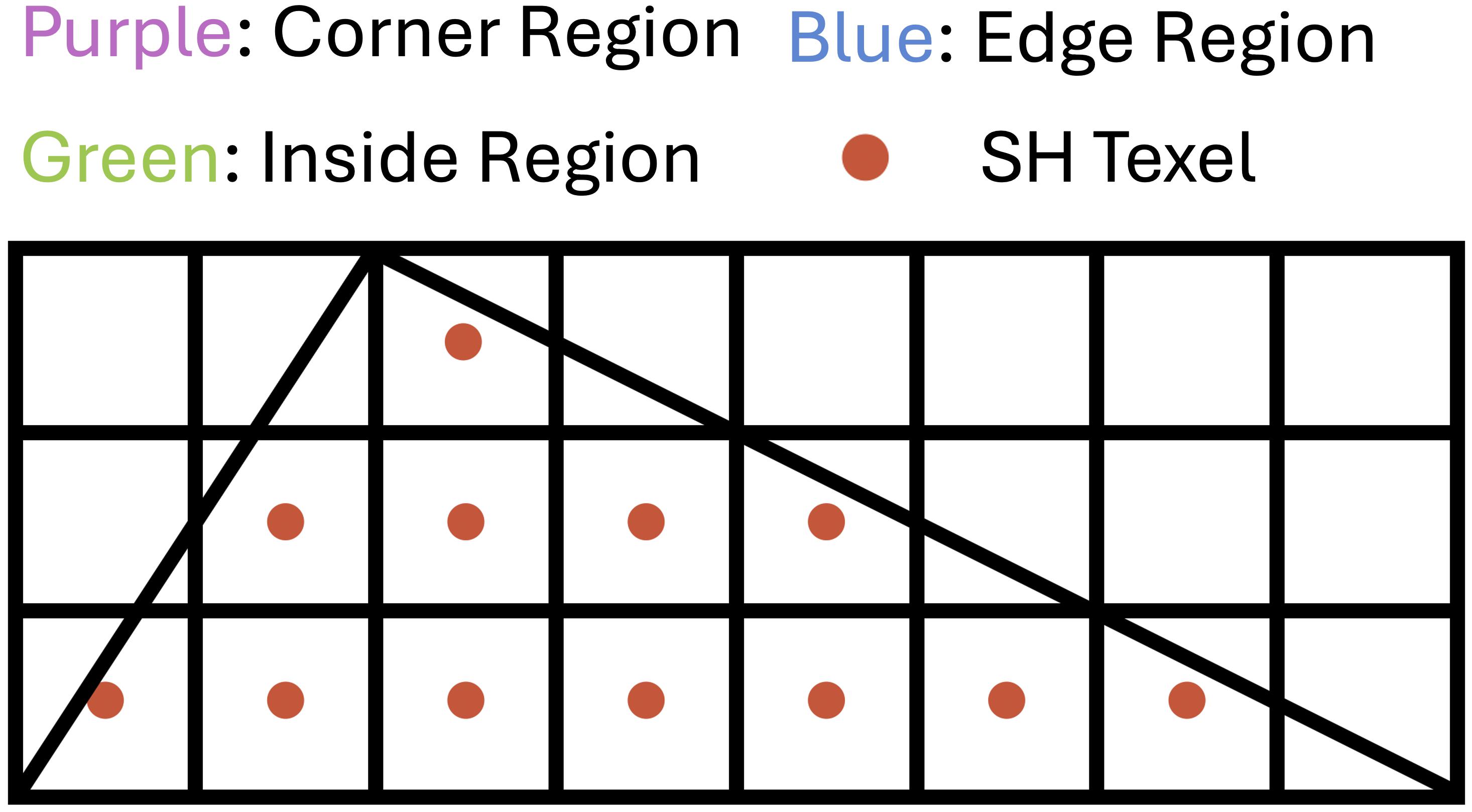}
        \label{fig:sh_density_arrange}
    }
    \hspace{10px}
    \subfloat[]{\includegraphics[width=0.3\linewidth]{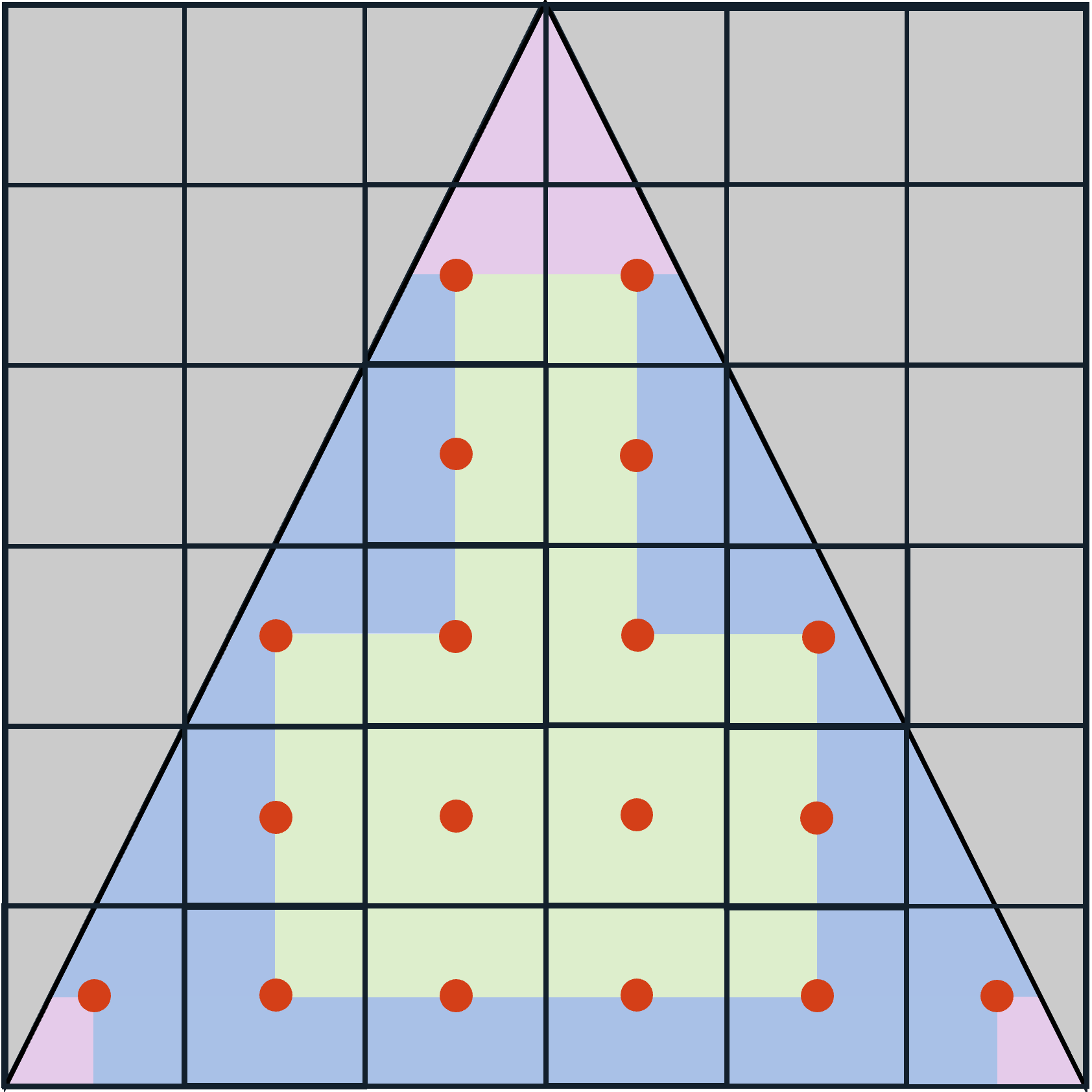}
    \label{fig:sh_3_regions}
    }
    
	\caption{Illustration of \textit{SH Texture's} resolution calculation and interpolation areas. Each mesh is allocated a dedicated \textit{SH Texture} whose resolution is calculated based on \textit{SH Density} and triangle's shape and size. For example, we calculate a SH Texture's resolution on an obtuse triangle, as in (a). We find the longest edge as the triangle base, and \textit{SH Texture}'s width/height is determined by dividing the triangle base/height length by the \textit{SH Density}. The \textit{SH Texels} whose centers lie outside the triangle are deemed invalid. (b): We define three regions for a \textit{SH Texture} to realize different interpolation strategies. \textit{Inside} (green): valid area where any sample point can find four texels for bilinear interpolation; \textit{Corner} (purple): valid area where sample points are above the highest valid \textit{SH Texel} center or on the lower-left side of the leftmost valid \textit{SH Texel} center, or on the lower-right side of the rightmost valid \textit{SH Texel} center. \textit{Edge} (blue): the rest valid area.}
\end{figure}


To address this, each mesh has to be assigned with different texture resolutions according to their distinctive shape and size. We calculate them by using \textit{SH Density}. \textit{SH Density} is defined as: \textit{the distance between two horizontally or vertically adjacent SH Texels in a texture, measured in meters.} It can be defined in meters because each \textit{SH Texel} within the \textit{SH Texture} can be assigned a 3D coordinate in world space. To determine texture resolutions, as shown in Fig. \ref{fig:sh_density_arrange}, we first find the longest edge of a triangle mesh and consider it as the triangle base. Next, we calculate the height relative to this base. Given that we know the world space coordinates of the triangle vertices, we can compute the length of the triangle base and height in world space, expressed in meters.
The arrangement of \textit{SH Texture} is aligned with its related mesh: the base of the \textit{SH Texture} coincides with the base of the triangle, and the height of the \textit{SH Texture} aligns with the direction of the triangle's height. To calculate the horizontal resolution of the texture, we divide the triangle's base length by the \textit{SH Density} to determine the number of texels that can be arranged along the triangle base, in other words, the resolution of the \textit{SH Texture} in the horizontal direction. Similarly, we divide the triangle's height by the \textit{SH Density} to obtain the number of \textit{SH Texels} that can be accommodated in the vertical direction, corresponding to the texture's vertical resolution. It is evident that, since the texture is rectangular when aligned with a triangle, there will inevitably be texels whose centers lie outside of the triangle boundary. We mark these texels as invalid, and they do not participate in subsequent training.

\begin{figure}[ht] 
	\centering
    \subfloat[]{
    \includegraphics[width=0.3\linewidth]{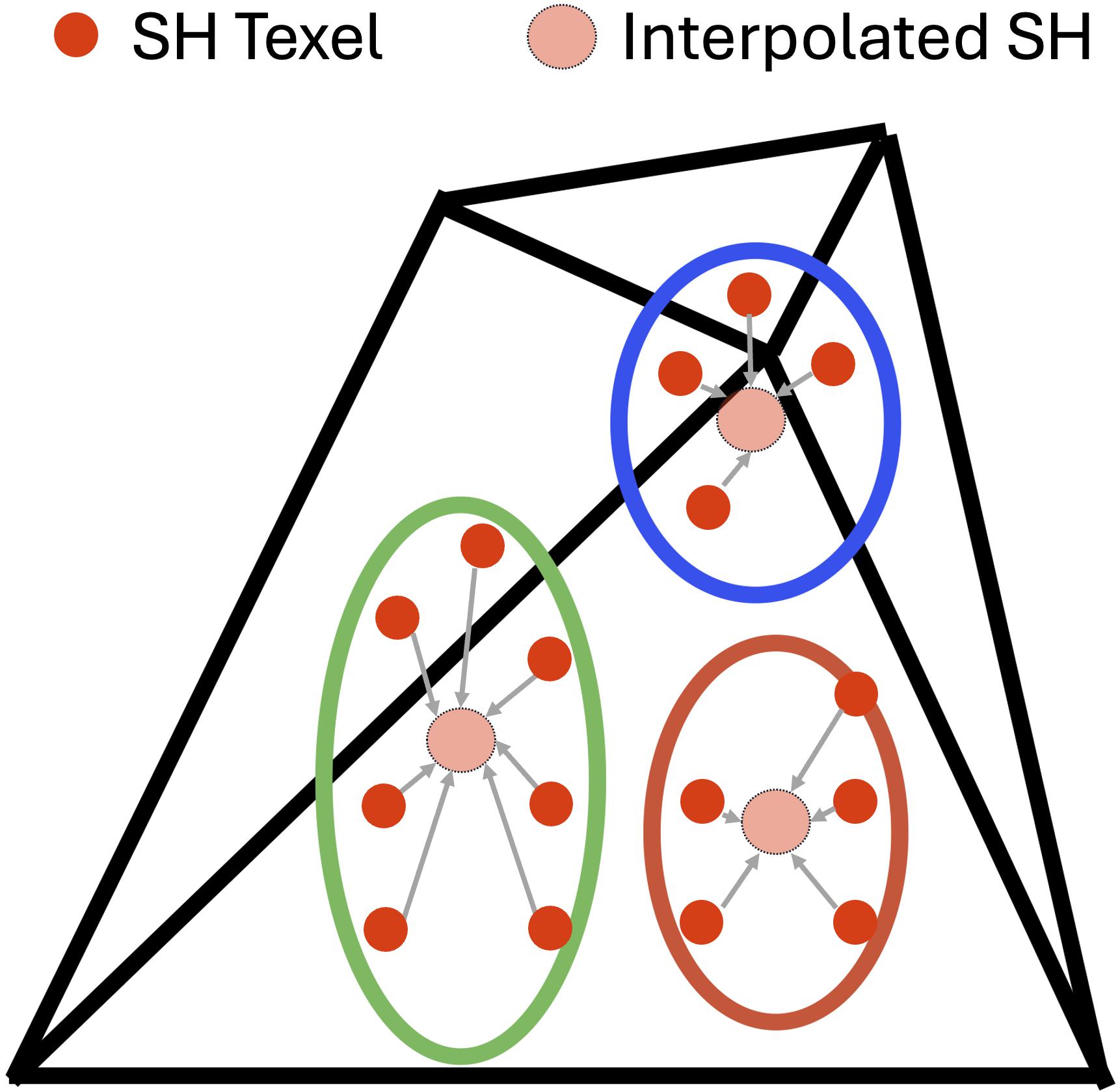}
    \label{fig:EWA_range_lerp_legend}
    }
    \subfloat[]{
    \includegraphics[width=0.3\textwidth]{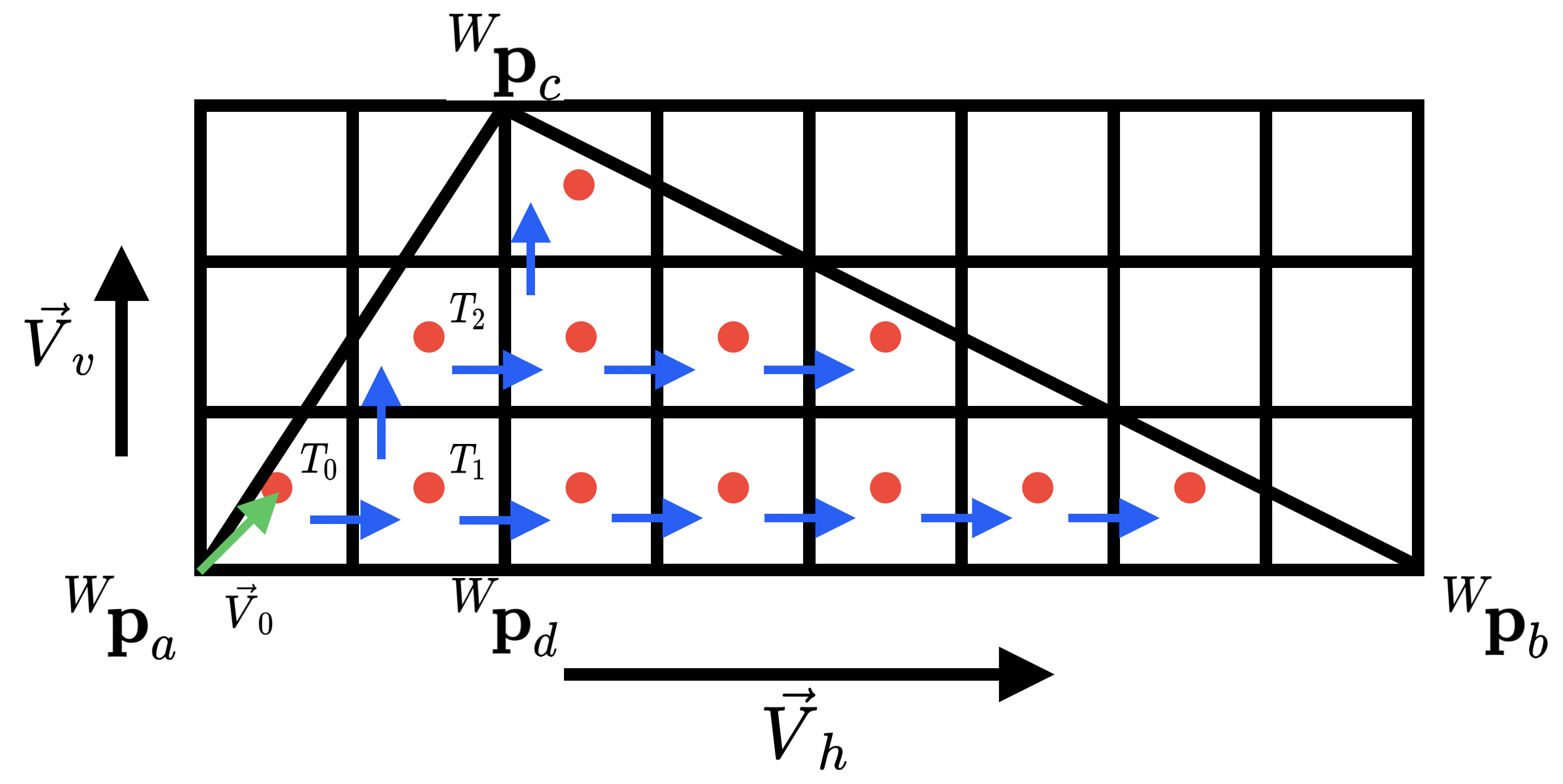}
    \label{fig:SH_world_position_calc}
    }




    
    

	\caption{Illustration of world space EWA and the calculation of world space coordinates of \textit{SH Texels}. (a):  In the \textit{Inside} region (red circle), we evaluate texels in the current texture; in the \textit{Corner} region (blue circle), we evaluate texels in neighbor textures that share the same triangle vertex; in the \textit{Edge} region (green circle), we evaluate \textit{SH Texels} in nearby \textit{SH Textures} that share the same edge. (b): The calculation of world space coordinates for \textit{SH Texels}.}
    \vspace{-0.2cm}
\end{figure}

Next, we calculate the world coordinates of each valid \textit{SH Texel} in \textit{SH Textures}. As illustrated in Fig. \ref{fig:SH_world_position_calc}, red dots represent valid \textit{SH Texel} centers, while blue lines indicate the calculation steps. We sequentially traverse and compute the world coordinates of all valid \textit{SH Texels} in the bottom row, moving to the row above and continuing this process until we reach the top row. 
This calculation is feasible because we know the world coordinates of the triangle vertices. We also know the direction and distance from the triangle vertices to the \textit{SH Texels} in world space, as well as the distances between every two adjacent \textit{SH Texels} (i.e. \textit{SH Density}). We begin at the bottom-left corner. First, we add the initial step vector $\mathbf{V}_0$ on the bottom-left triangle vertex ${}^W \! \mathbf p_a$ to obtain the world coordinate of the bottom-left-most \textit{SH Texel} $T_0$; then by adding the horizontal update vector $\mathbf{V}_h$ to $T_0$, we can find the world coordinates of its right-hand-side adjacent \textit{SH Texel} $T_1$. Similarly, we apply the vertical update vector $\mathbf{V}_v$ on $T_1$ to get the world coordinate of $T_2$. 
We denote the triangle vertices as ${}^W \! \mathbf p_a$, ${}^W \! \mathbf p_b$, ${}^W \! \mathbf p_c$, and ${}^W \! \mathbf p_d$ is the triangle foot.  The distance between two vertically or horizontally adjacent texels is the \textit{SH Density}, denoted as $\mathcal{D}$. So the horizontal $\mathbf{V}_h$ and vertical update vector $\mathbf{V}_v$ are calculated as follows: 
$$
\mathbf{V}_h =\frac{\mathbf{{}^W\mathbf p_b - {}^W\mathbf p_a}}{\|\mathbf{{}^W\mathbf p_b - {}^W\mathbf p_a}\|}  \mathcal{D}, ~~ \mathbf{V}_v =\frac{\mathbf{{}^W\mathbf p_c - {}^W\mathbf p_d}}{\|\mathbf{{}^W\mathbf p_c - {}^W\mathbf p_d}\|} \mathcal{D}
$$
Lastly, the initial step vector is: $\mathbf{V}_0 = {}^W \! \mathbf p_a + \frac{\mathbf{V}_h}{2} + \frac{\mathbf{V}_v}{2} $. We repeatedly add updated vectors to get all valid \textit{SH Texel}s' world coordinates.

\subsection{Deferred rendering}
\label{III.B}

  After initialization, we first perform deferred rendering at the beginning of each training step. This stage is not for rendering color, but for acquiring meshes in the current frustum and the information related to pixels and \textit{SH Texels}. The inputs for this stage include vertices, camera poses, and dummy textures. The outputs consist of sampling texture coordinates, EWA's ellipse coefficients, level of detail (LOD) level, view direction, and the world space coordinates at each pixel's sampling point (to be illustrated later). To obtain sampling texture coordinates of each pixel, we do not send all actual \textit{SH Textures} of all meshes to the GPU; instead, we use dummy textures.
  This is based on the facts that: 1) The combined memory footprint of all textures in the scene is too large to fit into GPU memory. 2) We only need texture coordinates at this stage, not sampling on textures. 3) Many textures share the same resolution configuration. 4) The number of distinct resolution configurations is orders of magnitude smaller than the total number of \textit{SH Textures}.
  To obtain the view directions and the world space coordinates of the sampling points, we employ barycentric interpolation during rasterization to interpolate the corresponding attributes of the triangle vertices. To get EWA's ellipse coefficients, we retrieve screen space pixel gradients and compute the ellipse coefficient in shader as outlined in \cite{EWA_original}. We also obtain the LOD level of each pixel in shader. All results are passed OpenGL's depth test and saved to offscreen GPU textures. 

\subsection{CUDA Renderer}
\label{III.C}
  We use CUDA-OpenGL interoperation to directly access the offscreen textures created by the deferred rendering stage on the GPU side. Based on that, for each pixel, we can identify which \textit{SH Texture} is related to and the associated sampling texture coordinates. So, we can identify which region of the \textit{SH Texture} this pixel belongs to (Sec. \ref{III.D}). Following this, we determine whether to use our hybrid interpolation or world space EWA filter to render based on each pixel's LOD level. For example, if the LOD level is greater than 1, we activate world space EWA; otherwise, we use hybrid interpolation.

\subsection{Hybrid SH Interpolation}
\label{III.D}

We have texture coordinates at each pixel's location, but relying only on bilinear interpolation on \textit{SH Texels} is not enough, especially when sampling at texture edges and corners. The bilinear method may not suffice as it requires four texels, which might not be available in this area. So it's natural to attend to adjacent \textit{SH Textures}.

First, we define the \textit{Valid Area}: as illustrated in Sec. 
\ref{III.A}, a valid \textit{SH Texel} lies within the scope of the triangle mesh. Similarly, a valid area is where sampling points are located within the triangle mesh's scope.
Then, we divide a \textit{SH Texture} into three regions: 
 \textit{Inside}, \textit{Corner}, and \textit{Edge} (as shown in Fig. \ref{fig:sh_3_regions}). \textit{Inside}: a valid area where we find four valid \textit{SH Texels} surrounding a sample point to conduct bilinear interpolation; 
\textit{Corner}: a valid area that is above the highest valid \textit{SH Texel} center, or located on the lower-left side of the left-bottom-most valid \textit{SH Texel} center, or the lower-right side of the right-bottom-most valid \textit{SH Texel} center; \textit{Edge}: the remaining valid area.

\HL{We implement different interpolation strategies in different regions.} For a sampling point in the \textit{Inside} region, since we can find nearby four texels, we use bilinear interpolation. 
If it's in the \textit{Edge} region, we cannot find four texels in self \textit{SH Texture} for bilinear interpolation. The possible number to find can be one, two, or three, depending on the location of the sampled texture coordinate within this region. We then locate one additional nearest \textit{SH Texel} in the neighbour \textit{SH Texture} who shares the same edge and use Inverse Distance Weighting (IDW) to interpolate among selected texels:
\begin{equation}
\begin{aligned}
&u(\mathbf{x})= \begin{cases}\frac{\sum_{i=1}^N w_i(\mathbf{x}) u_i}{\sum_{i=1}^N w_i(\mathbf{x})}, & \text { if } d\left(\mathbf{x}, \mathbf{x}_i\right) \neq 0 \text { for all } i \\ u_i, & \text { if } d\left(\mathbf{x}, \mathbf{x}_i\right)=0 \text { for some } i\end{cases}\\
&\text {where}\\
&w_i(\mathbf{x})=\frac{1}{d\left(\mathbf{x}, \mathbf{x}_i\right)^{0.9}}
\end{aligned}
\end{equation}
\HL{where $\mathbf{x}$ represents the sampled location, $\mathbf{x}_i$ denotes the locations of selected SH Texels, and $\mathbf{u}_i$ indicates the corresponding SH coefficients.} The distance function $d(\mathbf{x}, \mathbf{x}_i)$ represents the Euclidean distance in world space, $N$ indicates the number of \textit{SH Texels} to be interpolated.
If the sampled texture coordinate is in the \textit{Corner} region, plus the neighboring \textit{SH Texels} in self \textit{SH Texture},
we locate one nearest \textit{SH Texel} in \textbf{each} neighboring \textit{SH Texture} and apply IDW to interpolate among these texels.
We can use IDW because each \textit{SH Texel} has been assigned a world space coordinate during initialization, and the world space coordinate of the texture sample point (i.e., interpolation center of IDW) can be obtained from deferred rendering, as we mentioned in (Sec. \ref{III.B}). 
We choose IDW because the computational complexity is relatively low, which supports a large amount of pixel calculations while yielding acceptable interpolation results. 


\subsection{World Space EWA Filtering}
\label{III.E}
The results from the aforementioned method are quite commendable when cameras capture close-up scenery. However, a noticeable aliasing artifact remains when cameras face distant landscapes. 
To mitigate this, people often use MipMap \cite{mipmap} or other filtering methods. Considering the variability in the shape and size of meshes, the diverse camera viewpoints, and aiming to leverage \textit{SH Texels} on neighboring meshes, we propose using anisotropic filtering in world space. We extend EWA filtering \cite{EWA_original} from local texture space to world space.
This allows us to evaluate whether \textit{SH Texels} in the current \textit{SH Texture} and adjacent \textit{SH Textures} are within the ellipse range.
We then use in-range \textit{SH Texels} to calculate the elliptical weighted average.

The EWA's ellipse coefficients are calculated during the deferred rendering stage; they are originally in local texture space and can be transformed into world space. This is based on the following: 1) we have world space coordinates of the sampled point (i.e., the ellipse center), obtained during the deferred rendering stage. 2) we can calculate the ellipse's axis lengths in world space by multiplying the texel distance by the \textit{SH Density}. To determine whether the \textit{SH Texels} on neighbouring or self-textures lie within the ellipse in world space, we compute the world coordinates of the \textit{SH Texels}, which were determined in Sec. \ref{III.A}. Noted that, we only take adjacent \textit{SH Textures} into consideration (i.e., those who shares the edge or the vertex with current \textit{SH Texture}) for moderate computation overhead.

So far, since both the ellipse equation and \textit{SH Texels}' positions are in the same world space, we can determine if a \textit{SH Texel} on the current \textit{SH Texture} or its neighbouring \textit{SH Textures} falls within the ellipse's range by substituting the coordinates of the \textit{SH Texel} into the ellipse equation. Note that this algorithm assumes that the current mesh and adjacent meshes lie on the same plane. However, this assumption may not always be valid. Therefore, we compare the normals of the current and adjacent meshes; if the angle between the two normals exceeds $15$ degrees, we skip EWA for that adjacent mesh.
The strategy for selecting neighbouring texels is similar to that of the hybrid interpolation method: when the \textit{SH Texture} sample point is in the \textit{Inside} region, we apply EWA solely to the current texture; if in the \textit{Edge} region, we implement EWA on both the current texture and the neighbouring texture that shares the same edge; and if in the \textit{Corner} region, EWA is applied to the current texture and the neighbouring textures that share the same vertex.

\subsection{Adaptive SH Density Strategy}
\label{III.F}

Although the resolutions of \textit{SH Textures} differ due to varying mesh sizes and shapes, their \textit{SH Densities} remain the same. It is not ideal to set \textit{SH Density} globally constant, as the complexity of scene elements is inconsistent across scene regions. If \textit{SH Density} is too low, high-frequency details cannot be adequately represented through interpolation between sparse SHs. Conversely, setting \textit{SH Density} too high results in aliasing, as it results in the \textit{SH Density}, particularly in distant scenes, exceeding half the camera sampling rate (according to the Nyquist-Shannon Sampling Theorem \cite{sampling_theory}), leading to aliasing artifacts.
\begin{algorithm}[]
    \caption{\footnotesize Adaptive SH Density Strategy}
    \label{alg:adapDen}
    \footnotesize
    \SetKwInput{Not}{Notation}
    \Not{Initial SH Density $\rho_0$, number of converged mesh $N_{cvg}$, density update step $d$, threshold of max patiance number $T_{patiance}$, total number of meshes $N_{total}$.} \label{alg:ast:init}
    \KwIn{$-$ Current round PSNR average ${\mathcal{A}_i}$, PSNR variance ${\mathcal{V}_i}$, and retry times $C_i$; \newline $-$ Last round PSNR average ${{\bar{\mathcal{A}}_i}}$, PSNR variance ${{\bar{\mathcal{V}}_i}}$, and \textit{SH Density} ${\bar{\rho_i}}$. \newline $-$ Mesh set $\mathcal{M}_i$ and termination percentage of converged mesh $\epsilon_{T}$. }
    \KwOut{new \textit{SH Densities} $\rho_i$.}

    $N_{cvg}:=0, \rho_i:=\rho_0, C_i=0$\\ \label{alg:ast:init}
    \While{$\frac{N_{cvg}}{N_{total}} < \epsilon_{T}$}{
        \If{$\mathcal{M}_i$ is \textbf{Converged}}{\textbf{continue}}
        \vspace{0.1cm} 
        \If{$\mathcal{M}_i$ is first time update density}
        {
            $\rho_i:=\bar{\rho}_i + d$\\
            \textit{Recalculate SH Texture's resolution}
        }
        \vspace{0.1cm} 
        
            \If{${\mathcal{A}_i} > {{\bar{\mathcal{A}}_i}}$ \textbf{OR} ${\mathcal{V}_i} < {{\bar{\mathcal{V}}_i}}$}
            {
                $\rho_i:=\bar{\rho}_i + d$\\
                \textit{Recalculate SH Texture's resolution}
                }
            \Else{
            \If{$C_i < T_{patiance}$}
            {
                $C_i = \bar{C}_i + 1$
                \textbf{continue}
            }
            $\rho_i:=\bar{\rho}_i$ \\
            $N_{cvg} = N_{cvg} + 1$ \\
            $\mathcal{M}_i$ \textbf{Set Converged}
            }
    }
\end{algorithm}

Inspired by \cite{reduce_3dgs}, we propose an adaptive strategy to address this issue: we assign each mesh a globally low initial \textit{SH Density} and adjust the \textit{SH Density} changing direction (i.e., increase its density) through a metric method, aiming to adjust each \textit{SH Density} for improved metrics until no further enhancement is achievable. Specifically, after the initial \textit{SH Density} assignment, we train the first round to converge and then evaluate the averages and variances of PSNR based on all input views for each mesh separately.
Subsequently, we update \textit{SH Densities} according to each mesh's initial density update direction, recalculate \textit{SH Texture} resolutions, and begin a second round of training.
Once it converges, we can obtain another set of metrics (i.e., average and variance of PSNR) of each mesh. We compare these metrics with the previous counterparts. If the average PSNR of a mesh increases or the variance of a mesh's PSNR decreases, we consider the current update direction to be correct. Otherwise, we consider there's no need for further increasing the \textit{SH Density}.
Considering the stochastic nature of the training process, we introduce a ``patience'' parameter. If the metric does not improve, we allow the mesh another opportunity to optimize in the next training round. Should a mesh's metrics still not improve under these circumstances, we consider it converged, as it can no longer be optimized to enhance its metrics.
In Algorithm \ref{alg:adapDen}, $\mathcal{M}_i$ refers to a single mesh within the set $\mathcal{M}$ (i.e., all the meshes in a scene), and similarly, $C_i, \mathcal{A}_i, \mathcal{V}_i, \bar{\mathcal{A}}_i, \bar{\mathcal{V}}_i, \rho_i, \bar{\rho_i}$ refers to an attribute of a single mesh.

\subsection{Optimization}
{To recover scene radiance,} we use smooth $L1$ Loss ${\mathcal{S}}_{L_1}(\cdot)$ \cite{fast-RCNN} between the ground-truth pixel color $\boldsymbol{C}_i$ and rendered color $\hat{\boldsymbol{C}}_i$, where $i$ is the index of pixels:

\begin{small}
\begin{equation}
{Loss} = \sum_{i} {\mathcal{S}}_{L_1}(\boldsymbol{C}_i - \hat{\boldsymbol{C}}_i),
\end{equation}
\end{small}
where:

\begin{small}
\begin{equation}
  {\mathcal{S}}_{L_1}(x) =
  \begin{cases}
    0.5x^2& \text{if } |x| < 1\\
    |x| - 0.5& \text{otherwise}
  \end{cases}
\end{equation}
\end{small}

\section{Experiment}
In this section, we test and compare the rendering outcome of our framework on interpolation and extrapolation sequences under synthetic (Replica \cite{replica19arxiv}) and real dataset (Fast-LIVO2 Dataset \cite{fast_livo2}).

\subsection{Experiments Settings}
\subsubsection{Baselines}
Our work is compared to other state-of-the-art methods. For NeRF-based methods, we include InstantNGP \cite{instant-ngp} and M2-Mapping \cite{m2-mapping}. For 3DGS-based methods, we select 3D Gaussian Splatting (appearance-oriented) \cite{kerbl3Dgaussians} and MonoGS (RGBD-based) \cite{monoGS}. For texture-based methods, we choose the latest work \cite{chao2024texturedgaussiansenhanced3d} that combines textures with 3DGS.

\subsubsection{Metrics}
To evaluate rendering quality, we use Peak Signal-to-Noise Ratio (PSNR), Structural Similarity Index (SSIM), and Learned Perceptual Image Patch Similarity (LPIPS) to compare the rendering results with the corresponding ground truth image.

\subsection{Replica Dataset}
\begin{figure}[!t]
    \centering
    \includegraphics[width=1.0\linewidth]{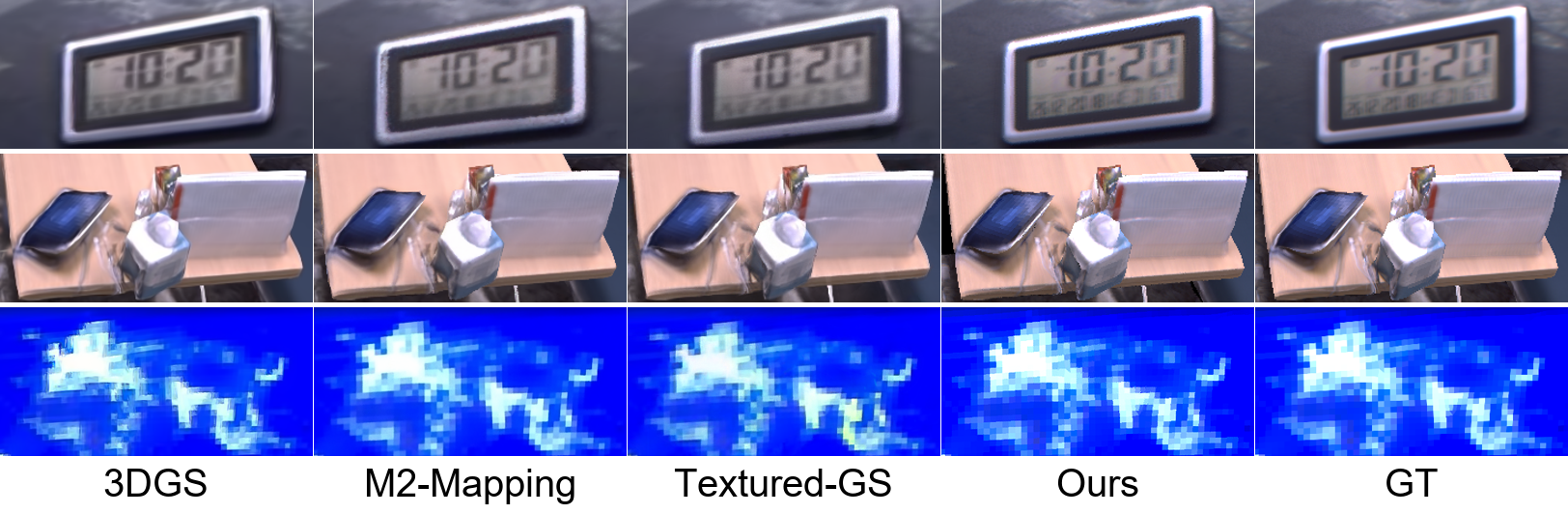}
    \caption{Comparison results on Replica Dataset's Office-0. The results demonstrate that our work captures more details in both extrapolation and interpolation sequences. In the first row, there's a zoomed-in view from an extrapolation perspective of a distant clock in the scene. In our results, the small numbers are clear enough to read, while the outcomes of other works are unreadable. In the second row's zoomed-in view from an extrapolation perspective, our result has sharper edges (blue pad) and more detail (white books). From an interpolation view in the last row, the pattern on the wall in our result is clearer and sharper compared to 3DGS, M2-Mapping and Textured-GS.} 
    \label{fig:replica}
\end{figure}

The Replica dataset \cite{replica19arxiv} is a synthetic dataset of indoor scenes. It generates images from specified viewpoints by rendering scene models using OpenGL. We employ the extrapolation and interpolation trajectories defined by M2-Mapping \cite{m2-mapping} for testing. As shown in Table \ref{tab:replica_result} and Figure \ref{fig:replica}, our proposed framework captures more precise details in both remote and nearby scenes, for both extrapolation and interpolation trajectories.

\input{table_I}

\begin{figure}[!t]
    \centering
    \vspace{0.3cm}
    \includegraphics[width=1.0\linewidth]{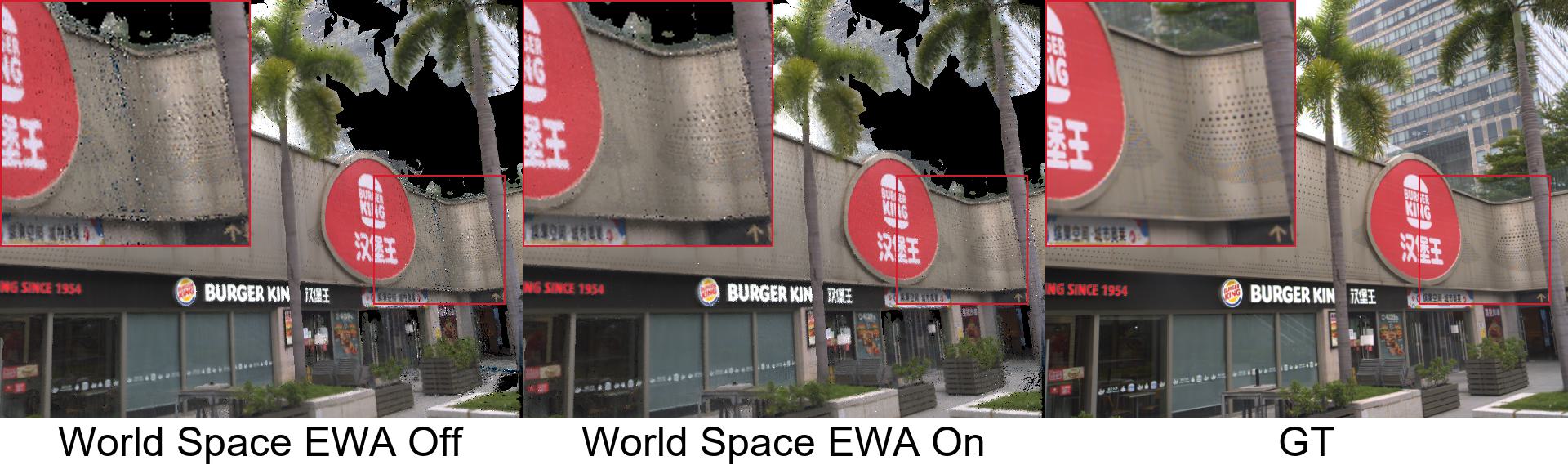}
    \caption{Comparison results between using and not using our proposed world space EWA filter. The noise here comes from an insufficient sampling rate, which results in an inadequate number of \textit{SH Texels} involved in the interpolation, causing the resulting colors to diverge from the GT.}
    \label{fig:EWA_compare}
\end{figure}

\begin{figure}[htbp]
    \centering
    \includegraphics[width=1.0\linewidth]{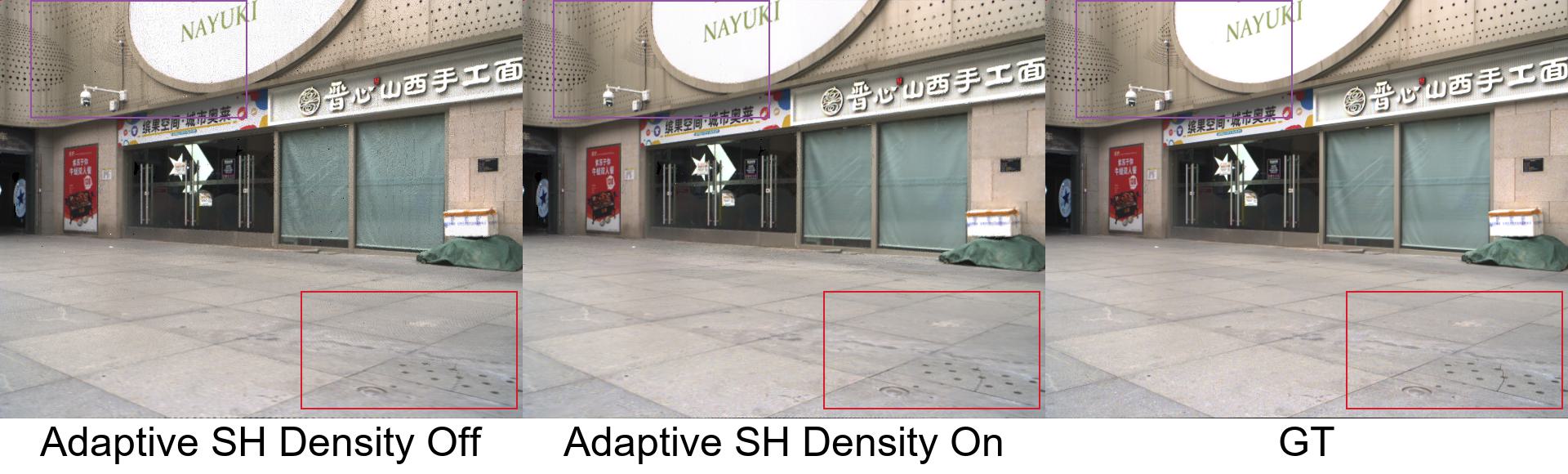}
    \caption{Comparison of our Adaptive SH Density Strategy.} 
    \label{fig:adaptive_den_compare}
\end{figure}

\begin{figure}[htbp]
    \centering
    \includegraphics[width=1.0\linewidth]{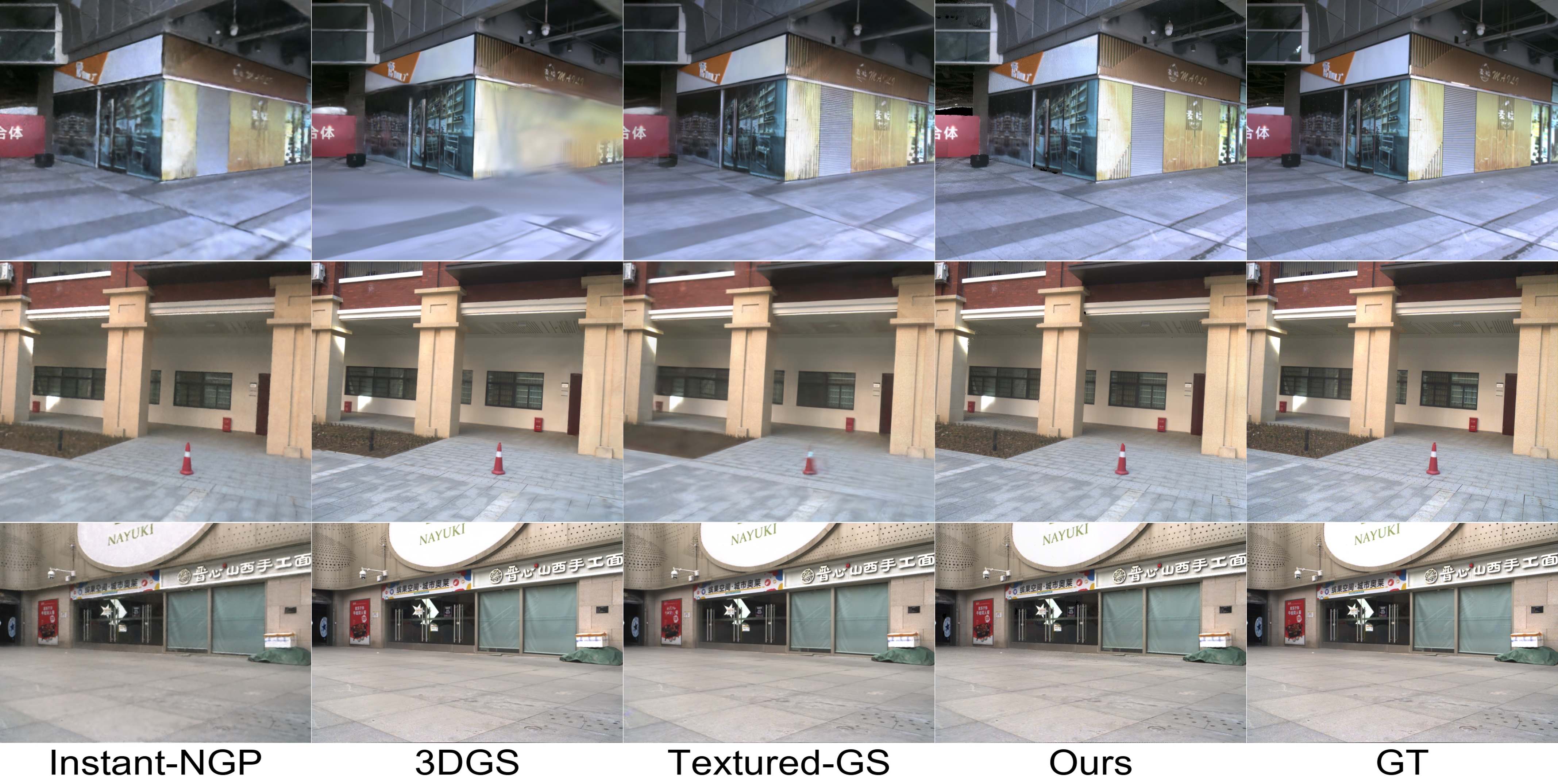}
    \caption{The qualitative results on FAST-LIVO2 Datasets.} 
    \label{fig:LIVO2_compare_all}
\end{figure}

\input{table_II}

\subsection{FAST-LIVO2 Dataset}
For real-world scenarios, we evaluate our framework on the FAST-LIVO2 datasets \cite{fast_livo2}. The localization results from FAST-LIVO2 are used as ground truth poses. To evaluate our algorithm's interpolation results, we split the pose-image-pair sequence generated by FAST-LIVO2 into the train and test set. Specifically, for a sequence, we extract every 8th pose-image-pair for the test set. The remains are for the train set. The quantitative results are shown in Tab. \ref{tab:fast_livo_result}. Our work shows promising rendering results compared to 3DGS \cite{kerbl3Dgaussians} and surpass the state-of-the-art in certain large-scale scenes. The qualitative results on the interpolation sequence are in Fig. \ref{fig:LIVO2_compare_all}. Our world space EWA filter can mitigate aliasing artifacts in distant view, in Fig. \ref{fig:EWA_compare}. What's more, we can render high-frequency regions well with high \textit{SH Density} while maintaining low \textit{SH Density} in mono color area, as shown in Fig. \ref{fig:adaptive_den_compare}.

\subsection{{Results}}
Rendering with hybrid interpolation alone on \textit{SH Textures} under fixed \textit{SH Density} does not yield satisfactory results due to aliasing artifacts caused by insufficient sampling rates. However, adopting the \textit{World Space EWA Filter} and \textit{Adaptive SH Density Strategy} can significantly improve the results.
As shown in Fig. \ref{fig:EWA_compare}, in the distant regions of the image, our world space EWA filter averages \textit{SH Texels} within the ellipse at each pixel's \textit{SH Texture} sampling point, resulting in smoother outcomes with less noise. We also implement the \textit{Adaptive SH Density Strategy} to further enhance our rendering quality, as illustrated in Fig. \ref{fig:adaptive_den_compare}.
Without it, rendering scenes with complex textures requires setting the global \textit{SH Density} very high. This leads to aliasing artifacts in distant regions where the \textit{SH Density} surpasses the camera's sampling rate (purple box). Similar artifacts, such as Moiré patterns, can also be observed within the red box. After implementing our adaptive strategy, we achieve a higher density in areas with complex textures (red box), which leads to improved rendering quality while keeping a lower \textit{SH Density} in areas with simple textures, thereby eliminating aliasing artifacts.

\section{Conclusion and Future Works}
\subsection{Conclusions}
In this work, we propose a 3D reconstruction and rendering framework called \textit{Mesh-Learner}, which is designed to learn view-dependent radiance while ensuring high compatibility with existing rasterization pipelines. We assign each mesh a \textit{SH Texture} whose resolution is calculated by \textit{SH Density}, and employ hybrid interpolation on \textit{SH Textures} at the sampling point of each pixel to obtain view-dependent color. 
We propose a World Space EWA Filter to reduce artifacts caused by insufficient sampling rates in remote scenes. Additionally, we implement an Adaptive SH Density strategy to determine the optimal \textit{SH Density} for each \textit{SH Texture}, further enhancing rendering quality while minimizing aliasing. Our work is evaluated on both synthetic (i.e., Replica \cite{replica19arxiv}) and real datasets (i.e., FAST-LIVO2 \cite{fast_livo2}), validating the effectiveness of our methods and demonstrating their potential for downstream tasks in robotics.

\subsection{Limitations and future works}
One primary limitation of our work is that we critically depend on the quality of the mesh. Specifically, in a scene with occlusion relationships, if the mesh of the foreground object is incomplete, it can lead to the colors that should belong to the foreground object being erroneously optimized onto the mesh of the background object in the training images. As a result, the incorrectly learned colors of the background object may be rendered in other viewpoints, producing erroneous outcomes. To address this, our future work will focus on optimizing mesh vertices with SH Textures to refine defective meshes and achieve better reconstruction and rendering results.

\FloatBarrier






{
\bibliographystyle{IEEEtran}
\bibliography{reference}
}

\end{document}

%% file: table_I.tex
\begin{table}[h]
    \centering
    \caption{Quantitative results on the FAST-LIVO2 dataset.}
    \label{tab:fast_livo_result}
    \resizebox{0.50\textwidth}{!}
    {
    \renewcommand{\arraystretch}{0.95} 
    \begin{tabular}{cccccccccc}
        \toprule
        \textbf{Metrics} & \textbf{Methods} & \textbf{Station} & \textbf{SYSU} & \textbf{CBD} &  \textbf{Avg.} \\
        \midrule
  
        \multirow{4}*{PSNR$\uparrow$}  
        & {InstantNGP} 
        & 24.111 & 22.517 & 22.468 & 23.720
        \\
        
        & {3DGS} 
        & \textbf{26.859} & \textbf{28.798} & 22.078 & \underline{25.825}
        \\

        & {Textured-GS}
        & \underline{25.458} & \underline{27.487} & \underline{24.625} & \textbf{25.856}
        \\

        & {Ours}
        & 25.453 & 26.010 & \textbf{25.409} & {25.624}
        \\
        \midrule
  
        \multirow{4}*{SSIM$\uparrow$}  
        & {InstantNGP} 
        & 0.780 & 0.792 & 0.779 & 0.783
        \\
  
        & {3DGS} 
        & \textbf{0.853} & \textbf{0.825} & {0.789} & \underline{0.822}
        \\

        & {Textured-GS} 
        & {0.804} & {0.787} & \underline{0.817} & {0.802}
        \\
        
        & {Ours}
        & \underline{0.833} & \underline{0.798} & \textbf{0.858} & \textbf{0.829}
        \\
        \midrule
  
        \multirow{4}*{LPIPS$\downarrow$}  
        
        & {InstantNGP} 
        & 0.258 & 0.367 & 0.357 & 0.327
        \\

        & {3DGS} 
        & \textbf{0.142} & \textbf{0.279} & {0.308} & \underline{0.243}
        \\

        & {Textured-GS} 
        & \underline{0.161} & {0.359} & \underline{0.246} & {0.255}
        \\
        
        & {Ours}
        & {0.172} & \underline{0.286} & \textbf{0.201} & \textbf{0.219}
        \\
        \bottomrule
    \end{tabular}
    }
    \vspace{-8pt}
\end{table}

%% file: table_II.tex
\begin{table*}[!ht] \small
\centering
\vspace{0.3cm}
\caption{Quantitative interpolation(I) and extrapolation(E) rendering results on the Replica Dataset. }
\label{tab:replica_result}
\resizebox{1.0\textwidth}{!}
{
\renewcommand{\arraystretch}{0.93} 
\begin{tabular}{ccccccccccc}
\toprule
\textbf{Metrics} & \textbf{Methods} & \textbf{Office-0} & \textbf{Office-1} & \textbf{Office-2} & \textbf{Office-3} & \textbf{Office-4} & \textbf{Room-0} & \textbf{Room-1} & \textbf{Room-2} & \textbf{Avg.} \\

\midrule
\multirow{6}{*}{SSIM(I)$\uparrow$}  & {InstantNGP} & 0.981                  & 0.980           & 0.963           & 0.960                  & 0.966           & 0.964           & 0.964           & 0.967           & 0.968            \\ 
                           & 3DGS       & \uline{0.987}          & 0.980           & \textbf{0.980}  & \uline{\textbf{0.976}} & \textbf{0.980}  & \textbf{0.977}  & \uline{0.982}   & \uline{0.980}   & \underline{0.980}            \\ 
                           & Textured-GS & 0.961 & 0.925 & 0.940 & 0.946 & 0.953 & 0.957 & 0.963 & 0.962 & 0.950\\
                           & MonoGS     & 0.985                  & 0.982           & \uline{0.973}   & 0.970                  & 0.976           & 0.971           & 0.976           & 0.977           & 0.975            \\ 
                           & M2-Mapping & 0.985                  & \uline{0.983}   & \uline{0.973}   & 0.970                  & \underline{0.977}           & 0.968           & 0.975           & 0.977           & 0.976            \\ 
                           & Ours       & \textbf{0.990}         & \textbf{0.986}  & \textbf{0.980}  & \uline{0.973}          & 0.976           & \uline{0.976}   & \textbf{0.988}  & \textbf{0.982}  & \textbf{0.981}            \\ 
\midrule
\multirow{6}{*}{PSNR(I)$\uparrow$}  & InstantNGP & 43.538                 & \uline{44.340}  & 38.273          & 37.603                 & 39.772          & 37.926          & 38.859          & 39.568          & 39.984           \\ 
                           & 3DGS       & 43.932                 & 43.237          & 39.182          & 38.564                 & \textbf{41.366} & \uline{39.081}  & \uline{41.288}  & \uline{41.431}  & 41.010           \\ 
                           & Textured-GS & 37.300 & 36.676 & 34.308 & 34.945 & 36.353 & 35.380 & 37.395 & 37.180 & 36.192\\
                           & MonoGS     & 43.648                 & 43.690          & 37.695          & 37.539                 & 40.224          & 37.779          & 39.563          & 40.134          & 40.034           \\ 
                           & M2-Mapping & \uline{44.369}         & \textbf{44.935} & \uline{39.652}  & \textbf{38.874}        & 41.318          & 38.541          & 40.775          & 40.705          & \uline{41.146}   \\ 
                           & Ours       & \textbf{44.689}        & 43.980          & \textbf{41.132} & \uline{38.723}         & \uline{41.352}  & \textbf{39.267} & \textbf{42.876} & \textbf{41.784} & \textbf{41.725}  \\ 
\midrule
\multirow{6}{*}{LPIPS(I)$\downarrow$} & InstantNGP & 0.046                  & 0.069           & 0.096           & 0.087                  & 0.089           & 0.079           & 0.094           & 0.094           & 0.082            \\ 
                           & 3DGS       & 0.040                  & 0.075           & 0.069           & 0.068                  & 0.065           & 0.060           & 0.057           & 0.075           & 0.064            \\ 
                           & Textured-GS & 0.066 & 0.160 & 0.138 & 0.085 & 0.094 & 0.059 & 0.066 & 0.080 & 0.093 \\
                           & MonoGS     & \uline{0.031}          & 0.048           & 0.061           & 0.057                  & 0.052           & \uline{0.055}   & \uline{0.050}   & 0.056           & 0.051            \\ 
                           & M2-Mapping & \uline{0.031}          & \uline{0.044}   & \uline{0.055}   & \uline{0.050}          & \underline{0.050}           & 0.062           & 0.056           & \uline{0.053}   & \uline{0.050}    \\ 
                           & Ours       & \uline{\textbf{0.025}} & \textbf{0.028}  & \textbf{0.027}  & \textbf{0.032}         & \textbf{0.036}  & \textbf{0.022}  & \textbf{0.023}  & \textbf{0.028}  & \textbf{0.028}   \\ 
\midrule
\multirow{6}{*}{SSIM(E)$\uparrow$}  & InstantNGP & 0.972                  & 0.961           & 0.934           & 0.938                  & 0.952           & 0.918           & 0.936           & 0.941           & 0.944            \\ 
                           & 3DGS       & 0.936                  & 0.897           & 0.924           & 0.917                  & 0.925           & 0.881           & 0.915           & 0.919           & 0.914            \\ 
                           & Textured-GS & 0.926 & 0.776 & 0.906 & 0.896 & 0.920 & 0.862 & 0.904 & 0.907 & 0.887 \\
                           & MonoGS     & 0.974                  & 0.961           & 0.945           & 0.942                  & 0.950           & 0.912           & 0.942           & 0.946           & 0.947            \\ 
                           & M2-Mapping & \uline{0.980}        & \uline{0.976}  & \uline{0.960}   & \uline{0.964}          & \uline{0.970}   & \uline{0.955}   & \uline{0.963}   & \uline{0.965}           & \uline{0.967}    \\ 
                           & Ours       & \textbf{0.987}         & \textbf{0.979}  & \textbf{0.973}  & \textbf{0.971}         & \textbf{0.978}  & \textbf{0.964}  & \textbf{0.975}  & \textbf{0.970}  & \textbf{0.974}   \\ 
\midrule
\multirow{6}{*}{PSNR(E)$\uparrow$}  & InstantNGP & 39.874                 & 39.120          & 31.274          & 32.135                 & 34.458          & 32.587          & 33.024          & 32.266          & 34.341           \\ 
                           & 3DGS       & 31.220                 & 29.959          & 27.411          & 26.442                 & 28.324          & 27.541          & 28.429          & 27.139          & 28.307           \\
                           & Textured-GS & 30.424 & 27.609 & 26.800 & 24.618 & 27.422 & 26.420 & 26.330 & 26.257 & 26.985  \\
                           & MonoGS     & 39.197                 & 38.818          & 29.740          & 29.664                 & 31.632          & 29.949          & 31.126          & 30.621          & 32.593           \\ 
                           & M2-Mapping & \uline{41.965}         & \textbf{42.215} & \uline{35.056}  & \textbf{37.465}        & \uline{38.667}  & \textbf{36.427} & \textbf{37.294} & \textbf{36.722} & \textbf{38.226}  \\ 
                           & Ours       & \textbf{42.480}        & \uline{41.751}  & \textbf{35.202} & \uline{36.154}         & \textbf{38.821} & \uline{35.510}  & \uline{36.872}  & \uline{36.531}  & \uline{37.915}   \\ 
\midrule
\multirow{6}{*}{LPIPS(E)$\downarrow$} & InstantNGP & 0.085                  & 0.117           & 0.176           & 0.185                  & 0.149           & 0.180           & 0.162           & 0.156           & 0.151            \\ 
                           & 3DGS       & 0.117                  & 0.177           & 0.167           & 0.181                  & 0.155           & 0.204           & 0.170           & 0.174           & 0.168            \\
                           & Textured-GS & 0.121 & 0.269 & 0.198 & 0.186 & 0.166 & 0.164 & 0.159 & 0.170 & 0.179 \\
                           & MonoGS     & 0.061                  & 0.104           & 0.136           & 0.137                  & 0.108           & 0.170           & 0.123           & 0.117           & 0.120            \\ 
                           & M2-Mapping & \uline{0.046}          & \uline{0.062}   & \uline{0.084}   & \uline{0.081}          & \uline{0.066}           & \uline{0.114}   & \uline{0.084}   & \uline{0.087}           & \uline{0.078}    \\ 
                           & Ours       & \textbf{0.027}         & \textbf{0.058}  & \textbf{0.061}  & \textbf{0.066}         & \textbf{0.051}  & \textbf{0.053}  & \textbf{0.046}  & \textbf{0.061}  & \textbf{0.047}   \\
\bottomrule
\end{tabular}
}
\end{table*}

%% file: root.bbl
\begin{thebibliography}{10}
\providecommand{\url}[1]{#1}
\csname url@samestyle\endcsname
\providecommand{\newblock}{\relax}
\providecommand{\bibinfo}[2]{#2}
\providecommand{\BIBentrySTDinterwordspacing}{\spaceskip=0pt\relax}
\providecommand{\BIBentryALTinterwordstretchfactor}{4}
\providecommand{\BIBentryALTinterwordspacing}{\spaceskip=\fontdimen2\font plus
\BIBentryALTinterwordstretchfactor\fontdimen3\font minus \fontdimen4\font\relax}
\providecommand{\BIBforeignlanguage}[2]{{%
\expandafter\ifx\csname l@#1\endcsname\relax
\typeout{** WARNING: IEEEtran.bst: No hyphenation pattern has been}%
\typeout{** loaded for the language `#1'. Using the pattern for}%
\typeout{** the default language instead.}%
\else
\language=\csname l@#1\endcsname
\fi
#2}}
\providecommand{\BIBdecl}{\relax}
\BIBdecl

\bibitem{mittal2023orbit}
M.~Mittal, C.~Yu, Q.~Yu, J.~Liu, N.~Rudin, D.~Hoeller, J.~L. Yuan, R.~Singh, Y.~Guo, H.~Mazhar, A.~Mandlekar, B.~Babich, G.~State, M.~Hutter, and A.~Garg, ``Orbit: A unified simulation framework for interactive robot learning environments,'' \emph{IEEE Robotics and Automation Letters}, vol.~8, no.~6, pp. 3740--3747, 2023.

\bibitem{imap}
E.~Sucar, S.~Liu, J.~Ortiz, and A.~Davison, ``{iMAP}: Implicit mapping and positioning in real-time,'' in \emph{Proceedings of the International Conference on Computer Vision ({ICCV})}, 2021.

\bibitem{meta_XR_SDK}
\BIBentryALTinterwordspacing
meta, ``Meta xr all-in-one sdk,'' 2025. [Online]. Available: \url{https://developers.meta.com/horizon/downloads/package/meta-xr-sdk-all-in-one-upm/}
\BIBentrySTDinterwordspacing

\bibitem{mvs_review}
\BIBentryALTinterwordspacing
X.~Yan, S.~Hu, Y.~Mao, Y.~Ye, and H.~Yu, ``Deep multi-view learning methods: A review,'' \emph{Neurocomputing}, vol. 448, pp. 106--129, 2021. [Online]. Available: \url{https://www.sciencedirect.com/science/article/pii/S0925231221004768}
\BIBentrySTDinterwordspacing

\bibitem{mildenhall2020nerf}
B.~Mildenhall, P.~P. Srinivasan, M.~Tancik, J.~T. Barron, R.~Ramamoorthi, and R.~Ng, ``Nerf: Representing scenes as neural radiance fields for view synthesis,'' in \emph{ECCV}, 2020.

\bibitem{kerbl3Dgaussians}
\BIBentryALTinterwordspacing
B.~Kerbl, G.~Kopanas, T.~Leimk{\"u}hler, and G.~Drettakis, ``3d gaussian splatting for real-time radiance field rendering,'' \emph{ACM Transactions on Graphics}, vol.~42, no.~4, July 2023. [Online]. Available: \url{https://repo-sam.inria.fr/fungraph/3d-gaussian-splatting/}
\BIBentrySTDinterwordspacing

\bibitem{nerfpp}
K.~Zhang, G.~Riegler, N.~Snavely, and V.~Koltun, ``Nerf++: Analyzing and improving neural radiance fields,'' \emph{arXiv:2010.07492}, 2020.

\bibitem{Yu2024MipSplatting}
Z.~Yu, A.~Chen, B.~Huang, T.~Sattler, and A.~Geiger, ``Mip-splatting: Alias-free 3d gaussian splatting,'' in \emph{Proceedings of the IEEE/CVF Conference on Computer Vision and Pattern Recognition (CVPR)}, June 2024, pp. 19\,447--19\,456.

\bibitem{reduce_3dgs}
\BIBentryALTinterwordspacing
P.~Papantonakis, G.~Kopanas, B.~Kerbl, A.~Lanvin, and G.~Drettakis, ``Reducing the memory footprint of 3d gaussian splatting,'' \emph{Proceedings of the ACM on Computer Graphics and Interactive Techniques}, vol.~7, no.~1, May 2024. [Online]. Available: \url{https://repo-sam.inria.fr/fungraph/reduced_3dgs/}
\BIBentrySTDinterwordspacing

\bibitem{3DGS_unity_addon}
\BIBentryALTinterwordspacing
aras p, ``Gaussian splatting playground in unity,'' 2023. [Online]. Available: \url{https://github.com/aras-p/UnityGaussianSplatting}
\BIBentrySTDinterwordspacing

\bibitem{unity_plugin_issue_artifacts}
\BIBentryALTinterwordspacing
K.~Ellersdorfer, ``Rendering artifacts close to the splats,'' 2024. [Online]. Available: \url{https://github.com/aras-p/UnityGaussianSplatting/issues/147/}
\BIBentrySTDinterwordspacing

\bibitem{unity_plugin_issue_err_occlusion}
\BIBentryALTinterwordspacing
JerryWu6288, ``The objects are not being rendered correctly,'' 2024. [Online]. Available: \url{https://github.com/aras-p/UnityGaussianSplatting/issues/99/}
\BIBentrySTDinterwordspacing

\bibitem{lin2023immesh}
J.~Lin, C.~Yuan, Y.~Cai, H.~Li, Y.~Zou, X.~Hong, and F.~Zhang, ``Immesh: An immediate lidar localization and meshing framework,'' \emph{arXiv preprint arXiv:2301.05206}, 2023.

\bibitem{fast_livo2}
C.~Zheng, W.~Xu, Z.~Zou, T.~Hua, C.~Yuan, D.~He, B.~Zhou, Z.~Liu, J.~Lin, F.~Zhu, Y.~Ren, R.~Wang, F.~Meng, and F.~Zhang, ``Fast-livo2: Fast, direct lidar–inertial–visual odometry,'' \emph{IEEE Transactions on Robotics}, vol.~41, pp. 326--346, 2025.

\bibitem{EWA_original}
N.~Greene and P.~S. Heckbert, ``Creating raster omnimax images from multiple perspective views using the elliptical weighted average filter,'' \emph{IEEE Computer Graphics and Applications}, vol.~6, no.~6, pp. 21--27, 1986.

\bibitem{pytorch}
\BIBentryALTinterwordspacing
A.~Paszke, S.~Gross, F.~Massa, A.~Lerer, J.~Bradbury, G.~Chanan, T.~Killeen, Z.~Lin, N.~Gimelshein, L.~Antiga, A.~Desmaison, A.~Kopf, E.~Yang, Z.~DeVito, M.~Raison, A.~Tejani, S.~Chilamkurthy, B.~Steiner, L.~Fang, J.~Bai, and S.~Chintala, ``Pytorch: An imperative style, high-performance deep learning library,'' in \emph{Advances in Neural Information Processing Systems}, H.~Wallach, H.~Larochelle, A.~Beygelzimer, F.~d\textquotesingle Alch\'{e}-Buc, E.~Fox, and R.~Garnett, Eds., vol.~32.\hskip 1em plus 0.5em minus 0.4em\relax Curran Associates, Inc., 2019. [Online]. Available: \url{https://proceedings.neurips.cc/paper_files/paper/2019/file/bdbca288fee7f92f2bfa9f7012727740-Paper.pdf}
\BIBentrySTDinterwordspacing

\bibitem{cuda}
\BIBentryALTinterwordspacing
NVIDIA, P.~Vingelmann, and F.~H. Fitzek, ``Cuda, release: 12.4.0,'' 2024. [Online]. Available: \url{https://developer.nvidia.com/cuda-toolkit}
\BIBentrySTDinterwordspacing

\bibitem{plenoxels}
{Sara Fridovich-Keil and Alex Yu}, M.~Tancik, Q.~Chen, B.~Recht, and A.~Kanazawa, ``Plenoxels: Radiance fields without neural networks,'' in \emph{CVPR}, 2022.

\bibitem{chao2024texturedgaussiansenhanced3d}
\BIBentryALTinterwordspacing
B.~Chao, H.-Y. Tseng, L.~Porzi, C.~Gao, T.~Li, Q.~Li, A.~Saraf, J.-B. Huang, J.~Kopf, G.~Wetzstein, and C.~Kim, ``Textured gaussians for enhanced 3d scene appearance modeling,'' 2024. [Online]. Available: \url{https://arxiv.org/abs/2411.18625}
\BIBentrySTDinterwordspacing

\bibitem{song2024hdgstextured2dgaussian}
\BIBentryALTinterwordspacing
Y.~Song, H.~Lin, J.~Lei, L.~Liu, and K.~Daniilidis, ``Hdgs: Textured 2d gaussian splatting for enhanced scene rendering,'' 2024. [Online]. Available: \url{https://arxiv.org/abs/2412.01823}
\BIBentrySTDinterwordspacing

\bibitem{Huang2DGS2024}
B.~Huang, Z.~Yu, A.~Chen, A.~Geiger, and S.~Gao, ``2d gaussian splatting for geometrically accurate radiance fields,'' in \emph{SIGGRAPH 2024 Conference Papers}.\hskip 1em plus 0.5em minus 0.4em\relax Association for Computing Machinery, 2024.

\bibitem{chen2022mobilenerf}
Z.~Chen, T.~Funkhouser, P.~Hedman, and A.~Tagliasacchi, ``Mobilenerf: Exploiting the polygon rasterization pipeline for efficient neural field rendering on mobile architectures,'' in \emph{The Conference on Computer Vision and Pattern Recognition (CVPR)}, 2023.

\bibitem{m2-mapping}
\BIBentryALTinterwordspacing
J.~Liu, C.~Zheng, Y.~Wan, B.~Wang, Y.~Cai, and F.~Zhang, ``Neural surface reconstruction and rendering for lidar-visual systems,'' 2024. [Online]. Available: \url{https://arxiv.org/abs/2409.05310}
\BIBentrySTDinterwordspacing

\bibitem{mipmap}
\BIBentryALTinterwordspacing
L.~Williams, ``Pyramidal parametrics,'' in \emph{Proceedings of the 10th Annual Conference on Computer Graphics and Interactive Techniques}, ser. SIGGRAPH '83.\hskip 1em plus 0.5em minus 0.4em\relax New York, NY, USA: Association for Computing Machinery, 1983, p. 1–11. [Online]. Available: \url{https://doi.org/10.1145/800059.801126}
\BIBentrySTDinterwordspacing

\bibitem{sampling_theory}
H.~Nyquist, ``Certain topics in telegraph transmission theory,'' \emph{Transactions of the American Institute of Electrical Engineers}, vol.~47, no.~2, pp. 617--644, 1928.

\bibitem{fast-RCNN}
R.~Girshick, ``Fast r-cnn,'' in \emph{2015 IEEE International Conference on Computer Vision (ICCV)}, 2015, pp. 1440--1448.

\bibitem{replica19arxiv}
J.~Straub, T.~Whelan, L.~Ma, Y.~Chen, E.~Wijmans, S.~Green, J.~J. Engel, R.~Mur-Artal, C.~Ren, S.~Verma, A.~Clarkson, M.~Yan, B.~Budge, Y.~Yan, X.~Pan, J.~Yon, Y.~Zou, K.~Leon, N.~Carter, J.~Briales, T.~Gillingham, E.~Mueggler, L.~Pesqueira, M.~Savva, D.~Batra, H.~M. Strasdat, R.~D. Nardi, M.~Goesele, S.~Lovegrove, and R.~Newcombe, ``The {R}eplica dataset: A digital replica of indoor spaces,'' \emph{arXiv preprint arXiv:1906.05797}, 2019.

\bibitem{instant-ngp}
\BIBentryALTinterwordspacing
T.~M\"uller, A.~Evans, C.~Schied, and A.~Keller, ``Instant neural graphics primitives with a multiresolution hash encoding,'' \emph{ACM Trans. Graph.}, vol.~41, no.~4, pp. 102:1--102:15, Jul. 2022. [Online]. Available: \url{https://doi.org/10.1145/3528223.3530127}
\BIBentrySTDinterwordspacing

\bibitem{monoGS}
H.~Matsuki, R.~Murai, P.~H.~J. Kelly, and A.~J. Davison, ``{G}aussian {S}platting {SLAM},'' in \emph{Proceedings of the IEEE/CVF Conference on Computer Vision and Pattern Recognition}, 2024.

\end{thebibliography}
